\documentclass{article}
\usepackage[utf8]{inputenc}

\usepackage{microtype}
\usepackage{graphicx}
\usepackage{subfigure}
\usepackage{booktabs} 
\usepackage{chngcntr}
\usepackage{amsmath}
\usepackage{xcolor}
\usepackage{amssymb}
\usepackage{hyperref}
\usepackage{stackengine}
\usepackage{authblk}
\usepackage[linesnumbered,lined,boxed]{algorithm2e}

\PassOptionsToPackage{numbers, compress}{natbib}

\usepackage[preprint]{neurips_2019}

\title{Object-oriented state editing for HRL}
\author{%
  \textbf{Victor Bapst, Alvaro Sanchez-Gonzalez, Omar Shams, Kimberly Stachenfeld,}\vspace{-1em}\\
  \textbf{Peter W.~Battaglia, Satinder Singh, Jessica B.~Hamrick}\\
  DeepMind, London, UK\\
  \texttt{vbapst@google.com}\\
}

\newcommand{\scenegraph}[0]{\mathcal{G}}
\newcommand{\connecttask}[0]{\texttt{connect}}
\newcommand{\filltask}[0]{\texttt{fill}}
\newcommand{\covertask}[0]{\texttt{cover}}
\newcommand{\editaction}[0]{\texttt{edit}}
\newcommand{\deleteaction}[0]{\texttt{delete}}
\newcommand{\addaction}[0]{\texttt{add}}

\begin{document}

\maketitle

\begin{abstract}
We introduce agents that use object-oriented reasoning to consider alternate states of the world in order to more quickly find solutions to problems.
Specifically, a hierarchical controller directs a low-level agent to behave \emph{as if} objects in the scene were added, deleted, or modified.
The actions taken by the controller are defined over a graph-based representation of the scene, with actions corresponding to adding, deleting, or editing the nodes of a graph.
We present preliminary results on three environments, demonstrating that our approach can achieve similar levels of reward as non-hierarchical agents, but with better data efficiency.
\end{abstract}

\section{Introduction}




Imagine giving advice to a nervous student about to give their first talk: only look at the people in the first row, and pretend there is no one else in the room.
By imagining the world \emph{as if} it were different, the student's nervousness subsides and they give a great talk.
This ability to entertain alternate states of the world is a key cognitive ability, beginning in childhood \cite{harris2000work} and supporting behavior ranging from everyday problem solving to scientific thought experiments \cite{clement1994use,gendler1998galileo,trickett2007if}.

Standard approaches in deep reinforcement learning have so far only touched on a small subset of possible ways to imagine the world being different.
For example, agents that use planning \cite{hamrick2019analogues} consider what might happen if an alternate action were taken.
Skill-based hierarchical agents such as \cite{heess2016learning,oh2017zero} choose between alternate behaviors via a high-level manager.
Hindsight-based agents such as \cite{andrychowicz2017hindsight,sahni2019visual} assume, in retrospect, that a goal is different than it actually was.
Goal-conditioned hierarchical agents such as \cite{kulkarni2016hierarchical,nair2018visual,vezhnevets2017feudal} create hypothetical goals for a low-level agent to achieve.
However, in these examples, alternate states of the world are often focused on intermediate goals that the agent must achieve as a prerequisite for completing the main task.
Yet, it can sometimes be useful to imagine \emph{analogous} rather than intermediate states of the world, enabling the transfer of behavior from one situation to another.
For example, what if we tried to work around an object that isn't actually there? Or, what if we acted as if an object functioned differently than it does? 
We suggest that such forms of counterfactual reasoning can support an alternate mechanism for directing and constraining hierarchical behavior than intermediate goals.

In our approach to hierarchical reasoning, we allow a high-level controller to modify the objects in the observations of a pretrained low-level agent, thus instructing the low-level agent to behave \emph{as if} the world were different.
These observations are encoded as a graph, with the nodes of the graph corresponding to objects in the scene.
The actions of the controller are then modifications of this graph.
More precisely, our controller can \editaction{} node properties, \deleteaction{} nodes (and their associated edges), or \addaction{} nodes (and associated edges) to the graph.
We put our idea to the test on a family of tasks inspired by the construction domain \cite{bapst2019construction}.
Each task comes in two versions, which we refer to as \emph{pretraining} scenes for training low-level skills, and \emph{transfer} scenes for further training and evaulation.

\begin{figure}[t!]
\begin{center}
    \includegraphics[width=\textwidth]{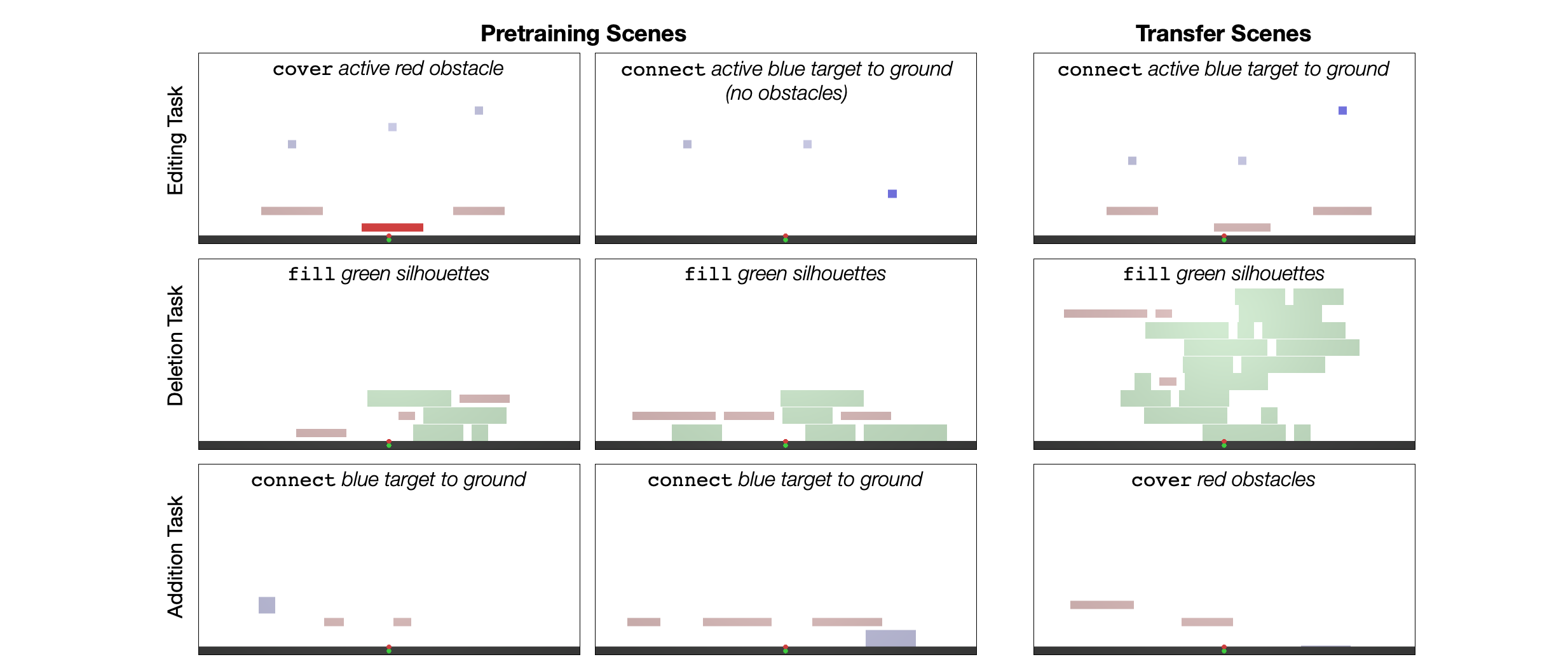}

    \caption{Pretraining (left) and transfer tasks (right). \textbf{First row:} Editing task. In pretraining, the goal is to \covertask{} or \connecttask{} an active obstacle or target (shown in a bright color), respectively. In the transfer task, the goal is to \connecttask{} an active target above an obstacle. \textbf{Second row:}  Deletion task. The agent must \filltask{} in the green areas while avoiding the red obstacles. The pretraining scenes consist of fewer targets at lower vertical positions than the transfer scenes. \textbf{Third row:} Addition task. In the pretraining task, the agent has to \connecttask{} the blue target. In the transfer task, it has to \covertask{} the red obstacles from above.}
    \label{fig:tasks_snapshots}
\end{center}
\end{figure}

\section{Hierarchical Agent Architecture}

Both the controller and low-level agent act on a fully-connected scene graph, $\mathcal{G}$, whose nodes correspond to the objects in the scene.
Each node has attributes associated with object properties, such as position, size, and object type.
All agents except the heuristic controller (described below) process $\mathcal{G}$ using a graph network \cite{battaglia2018relational} and output Q-values on the edges and/or nodes of the graph, which are trained using Q-learning.

The \textbf{low-level} agent's actions defined are on the edges of the graph $\scenegraph$ and correspond to placing blocks relatively to others, similar to \cite{bapst2019construction}. This agent is pretrained until convergence on the pretraining version of the tasks.
We use two architectures for our controller. In the first version (the \textbf{heuristic controller}), we use a non-learned, heuristic baseline, that we specifically hard-code for each of the tasks afterwards. In this case, we also explore finetuning the low-level policy $\pi$.
The second version (the \textbf{neural controller}) is a graph network which produces Q-values on \emph{either} the nodes or edges of $\mathcal{G}$ (see Sec.~\ref{sec:tasks} for details about the actions). Moreover, as our physical environment can be simulated, the same goes of the environment seen by the controller, and we can augment the latter with planning, as in \cite{bapst2019construction}. 
We additionally compare our results to the following \textbf{baselines}: directly applying the pretrained agent to the task without re-trainining, a model free agent directly trained on the transfer version of the task, and a model based agent directly trained on the transfer version of the task.

\section{Tasks}
\label{sec:tasks}

In order to test the capacity of our approach, we developed four tasks which are variations on the construction tasks from \cite{bapst2019construction}. The original tasks involved stacking blocks to achieve various objectives, such as to \filltask{} a silhouette shape with blocks, to \connecttask{} a target in the sky to the ground, or to \covertask{} other objects from above (without touching them). As with these original tasks, we allow our agent to use three different sized blocks which may be optionally made ``sticky'' (for a price), so that they stick to other objects upon touching them.

\begin{figure}[t]
    \centering
    \includegraphics[width=\textwidth]{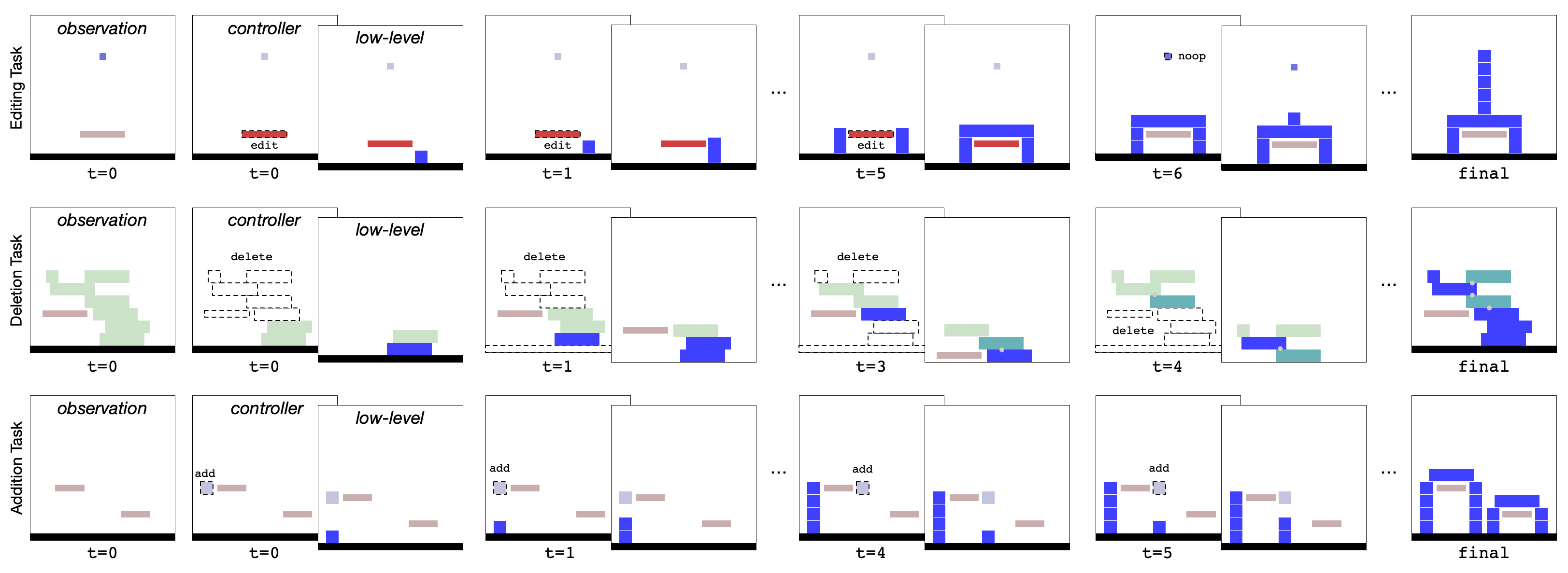}
    \caption{Example controller behavior. The controller takes in the true observation from the environment, and can modify it either by \emph{editing} an object to mark it as active, \emph{deleting} objects that are too far from a selected object, or \emph{adding} a new object. The modified observation is then passed to the low-level agent. See text for details.}
    \label{fig:controller}
\end{figure}

\paragraph{Editing task}

The scene contains between 1 and 3 obstacles at various heights, each with a target above it (Fig.~\ref{fig:tasks_snapshots}, top right).
The agent has to \connecttask{} one of these targets (marked as ``active'' with a special one-hot encoding in $\scenegraph$) to the ground by overlapping a block with it.
The low-level policy is pretrained on two tasks: either to \connecttask{} the active target in a scene without obstacles or to \covertask{} a given active obstacle from above.

The controller acts on the scene graph $\scenegraph$ by making an \editaction{} to which object is currently marked as activated. Its action is therefore attached to the nodes of the graph, and takes the form $\scenegraph \to v \in \textrm{nodes}(\scenegraph)$. The graph $\scenegraph'$ is then formed by making vertex $v$ the active vertex in $\scenegraph$. The heuristic controller makes the obstacle under the target active for a fixed number of steps (sufficient to cover the obstacle), and then activates the target to connect.

\paragraph{Deletion task}

In this task, the agent must \filltask{} between 10 and 20 silhouettes, on up to 10 levels (see Fig.~\ref{fig:tasks_snapshots}, middle row, right). 
The low-level agent is trained on easier scenes, consisting of at most 6 targets, on at most 3 horizontal levels. 
The reward is of +1 for each block which is sufficiently filled, while sticky blocks have a cost of -0.5.

The controller acts on the scene graph $\scenegraph$ by choosing to \deleteaction{} objects outside of a vertical band $[y_{\textrm{min}}, y_{\textrm{max}}]$. It does so by selecting a vertex $v$ in the graph; only the objects whose center $y$-coordinate is within $\pm 1.5$ block height of the selected object center's $y$-coordinate are kept (corresponding to 3 horizontal rows of objects); this is a form of hard attention on the graph. The $y$-coordinates of all the shown objects are also re-centered such that the bottom of the lowest  present object is at exactly zero. The heuristic controller works by selecting the lowest target which has not yet been connected.

\paragraph{Addition task}

In this task, the agent has to \covertask{} obstacles from above without touching them.
The low-level policy is pretrained on scenes from the same distribution, but it must \connecttask{} a target (which is placed to the side or on top of one of the obstacles) rather than \covertask{} obstacles.

The controller's modification of the scene from the transfer task is to \addaction{} a target block. Its action is of the form $\scenegraph \to (e, x) \in \textrm{edges}(\scenegraph) \times \{1, \Delta\}$, where the start vertex of the edge  $e$ indicates which block $u$ (from the bottom of the screen) should be used as a target block to match, the end vertex of $e$ indicates relatively to which block $v$ it should be placed, and $x$ selects the relative lateral positioning of $u$ with respect to $v$. The heuristic controller iterates over the obstacles from left to right, placing a small target block to the left of it until the agent connects it, then to the right of it until the agent connects it, and then a large target on top of it.

\paragraph{Combined}

Finally, we experiment with a task combining the three previous tasks.
The hierarchical agent is trained on the transfer versions of the three tasks presented above, with each episode comprising a different, randomly selected task.
The low-level agent is pretrained on a uniform mixture of the corresponding pretraining versions of the tasks.

The controller acts on the scene graph $\scenegraph$ as
$\scenegraph \to (v, k) \in \textrm{nodes}(\scenegraph) \times \{1, 2\}  \; \; \textrm{or} \;\; (e, x) \in \textrm{edges}(\scenegraph) \times \{1, \dots, \Delta\}$,
where $k$ is a node action type, and $v$, $e$ and $x$ are as above.
The action is then interpreted as follows: (1) If the action is a node action, and if $k = 0$, then an edit action is taken. (2) If the action is a node action, and if $k = 1$, then a deletion action is taken. (3) If the action is an edge action, \textit{and if the task is the object addition task}, then an object addition action is taken; otherwise this is a no-op.
We do not propose a heuristic controller for the combined case.

\section{Results}

\begin{figure}[t!]
\begin{center}
    \includegraphics[width=0.95\textwidth]{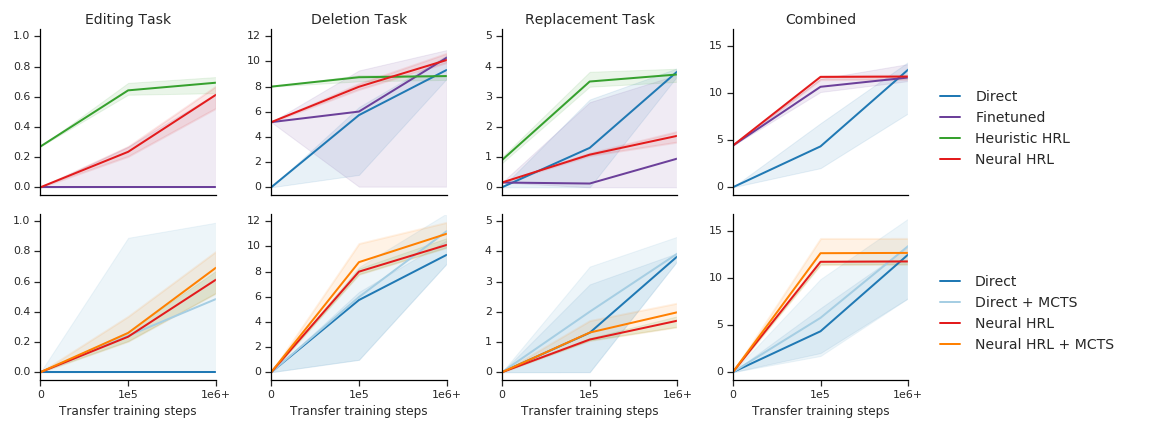}
    \caption{Agent reward as a function of the training budget. Lines show the median over 10 seeds, and shaded areas indicate best and worst seeds. The rightmost point correspond to converged agents.  We provide videos of our controller agents on the first three tasks at \url{https://tinyurl.com/y2o655pm}. }
    \label{fig:controller_results_3_tasks}
\end{center}
\end{figure}


On the top row of Fig.~\ref{fig:controller_results_3_tasks} we compare our HRL agent (red) to an agent directly trained on the task (blue), an HRL agent with a heuristic controller (green), and finetuning the pretrained agent (purple). Before training on the transfer scenes, only the heuristic controller obtains a non-trivial reward.

With a small training budget of $10^5$ learner steps (about $25\, 000$ actor steps) that it uses to finetune the low-level policy $\pi$, the performance of the heuristic HRL agent quickly improves and already almost reaches its large training budget performance. In this low data regime, the HRL agent (red) also quickly trains its controller and outperforms both an agent directly trained on the task (blue) but also the finetuned pretrained agent (purple) in 3 out of 4 tasks. This demonstrates that the hierarchical setup allows quicker skills reuse, both when the low-level policy or the high-level policy are trained, compared to directly (re-)training on the transfer task.

In the large training budget regime, the heuristic HRL agent suffers from the rigidity of the handcrafter controller and is outperformed by other approaches, particularly in the deletion task. On the other hand, the learned HRL controller benefits from this increased training budget and outperforms the finetuned agent in all the tasks. Only an agent directly trained on the task achieves better performance (except in the editing task), because it has the most flexibility to adapt to the transfer task.

We attempted to learn both the low-level policy and the controller at the same time, but did not manage to improve over our current learned controller in the HRL agent.
However, given the results with the heuristic controller, we believe that further investigation of the learning dynamics holds promise for simultaneously learning both the controller and finetuning the low-level policy.

Finally, on the second row of Fig.~\ref{fig:controller_results_3_tasks} we study the effect of adding planning to the controller component of the HRL agent. In both the HRL agent as well as the agent directly trained on the tasks, planning (here with a budget of 10) gives a slight improvement to the results.

\section{Conclusion}

We presented an agent capable of taking object-oriented actions to modify an observation before feeding it to a pretrained low-level agent.
These object-oriented actions allow the high-level controller to direct the low-level agent to behave \emph{as if} the world were different than it actually is, thus potentially enabling skill reuse in new scenarios.
Our hierarchical agent improves over the performance of the pretrained agent, and can even outperform an agent directly trained on the transfer tasks.
We suggest that such hierarchical object-oriented approaches hold promise in developing more efficient, flexible, and compositional agents going forward.

\clearpage
\bibliography{main}
\bibliographystyle{plain}

\clearpage
\appendix
\section{Appendix}

\subsection{Task details}

Our simulated task environment is a continuous 2D world implemented in
Unity \cite{juliani2018unity} with the Box2D physics engine
\cite{catto2013box2d}, based on those explored in \cite{bapst2019construction}.
Each episode contains unmoveable obstacles,
target objects, and floor, plus movable rectangular blocks
which can be picked up and placed.
On each step of an episode, the agent chooses an available
block (from below the floor), and places it in the scene
(above the floor) by specifying its position. There is an unlimited supply of blocks of each size, so the same block can be picked
up and placed multiple times. The agent may also attach
objects together by assigning the property of “stickiness”
to the block it is placing. Sticky objects form unbreakable,
nearly rigid bonds with objects they contact. After the agent places a block, the environment runs physics forward until all blocks come to rest.

An episode terminates when: (1) a movable block makes
contact with an obstacle, either because it is placed in an
overlapping location, or because they collide under physical
dynamics; (2) a maximum number of actions of 14 is exceeded;
or (3) the task-specific termination criterion is achieved
(described below). The episode always yields zero reward
when a movable block makes contact with an obstacle.

The full rendered scene spans a region a size of 16x16. At the beginning of the episode, the agent has access to 7 available blocks: three small, three medium and one large block (corresponding to respective widths of 0.7, 2.1 and 3.5, all with height 0.7). The physics simulation is run for 20 seconds after the agent places each block to make sure that the scene is at an equilibrium position before the score is evaluated, and before the agent can place the next block.

Each of the scene from the tasks below is procedurally generated, with task specific details given below.

\paragraph{Editing task}

The transfer scenes have one active target block per scene, indicated by a special one-hot encoding in the observation. The agent is given a reward of +1 if it touches the center of this block, and using glue has a cost of -2, so that no glue should be used on an optimal strategy. The episode ends when the active target block has been connected to the ground. The maximal reward that can be obtained is 1.

In the first pretraining task, there are between 1 and 3 targets at various heights (and no obstacles), and the goal is to connect the active target to the ground. In the second pretraining task, the distribution is identical to the transfer task distribution except for the reward function, which rewards covering the active obstacle (proportionally to its length) rather than connecting an object to the ground. The episode ends when more then 99\% of this obstacle length has been covered. The agent must recognize which of the two pretraining task it is solving from the observation $\scenegraph$.

\paragraph{Deletion task}

Each transfer scene is comprised of 10 to 20 targets and 0 to 6 obstacles, arranged in up to 10 layers, while the pretraining scenes contain between 3 and 6 targets on up to 3 layers with at most 3 obstacles. Levels are generated similarly to the procedure described in \cite{bapst2019construction}. The reward function is: +1 for each placed block which overlaps at least 90\% with a target block of the same size; and -0.5 for each block set as sticky. The maximal reward that can be obtained is upper-bounded by 15.

\paragraph{Addition task}

There are between 1 and 3 obstacles, at different heights. The agent receives a reward proportional to the length of obstacles being vertically covered, and the episode ends when all the obstacles have been covered. If an obstacle is touched, the episode ends and no further reward is given. Using glue gives a negative reward of -2, so that using it is effectively prohibited. The maximal reward that can be obtained is upper-bounded by 4.5.

\paragraph{Combined task}

The transfer task is a uniform combination of the three transfer tasks above, and the agent has no indication of which task it should be solving (but this can be recovered from $\scenegraph$). We monitor the rewards for each of the three sub-tasks, and do not normalize the total reward given to the agent. Pretraining is performed in the same way, with a uniform mixture of the three pretraining tasks. The pretrained agent obtains ceiling performance on the three pretraining sub-tasks.

\subsection{Agent details}

The episode loop of the agent is summarized on \autoref{algo:main_algo}.

\begin{algorithm}[H]
\SetAlgoLined
\SetKwInOut{Input}{input}\SetKwInOut{Output}{output}
\Input{High level policy $\Pi$, low level policy $\pi$, initial state $\scenegraph$ of new episode}
 $\textrm{buffer}_\Pi$ = $[]$;  $\textrm{buffer}_\pi$ = $[]$\;
 \While{$\scenegraph$ is not terminal}{
  take action $A$ according to $\Pi(\scenegraph)$; compute modified state $\scenegraph'$ according to $A$\;
  take action $a$ according to $\pi(\scenegraph')$; observe next state $\scenegraph''$ and reward $r$\;
  add $(\scenegraph, A, r)$ to $\textrm{buffer}_\Pi$; add $(\scenegraph', a, r)$ to $\textrm{buffer}_\pi$\;
  $\scenegraph \leftarrow \scenegraph''$;
 }
 \caption{The hierarchical agent episode loop. $\textrm{buffer}_\Pi$ and $\textrm{buffer}_\pi$ are then used to perform learning of $\Pi$ or finetuning og $\pi$.}
 \label{algo:main_algo}
\end{algorithm}

\paragraph{Network}

The controller's graph network consists of an encoder, followed by 3 steps of reasoning and a decoder, where we copied all the hyper-parameters from \cite{bapst2019construction}.

We use 15 discretization steps for the low level agent's action, and $\Delta=7$ discretization steps for the controller in the objects addition and combined tasks.

 \paragraph{Learning procedure}
 
 We perform Q-learning with a learning rate of $2\times 10^{-4}$, using the Adam optimizer. We use a batch size of 16 and a replay ratio of 4. We use a discount of $0.98$ when accumulating rewards for learning.

No curriculum is performed in any of the tasks. We use the same adaptive exploration schedule as was used in \cite{bapst2019construction}.

We use a distributed setup with up to 128 actors (for the largest MCTS budgets) and 1 learner. Our setup is synchronized to
keep the replay ratio constant.

 \paragraph{MCTS}
 
 We perform MCTS expansions similarily to \cite{bapst2019construction}, with a search budget of 10, a UCT constant of 2, and a sampling of the action in two stages (first selecting a node or an edge, and then the second dimension of the action).
 We also add an additional term to the loss function to encourage the Q-values output by the neural network to be more similar to those output by MCTS.
 Specifically, we add an cross-entropy loss term between the softmax of the Q-values output by the neural network and the softmax of the Q-values output by MCTS.
 We found this additional loss term helped to stabilize training and generally improved performance.

\paragraph{Evaluation details}

All our model free methods are trained for 10 million steps, and our model based methods were trained for 2 million steps. We report the best performance across 10 seeds over bins of $10^4$ consecutive learner steps during training.

\subsection{Results details}

\begin{figure}[h!]
\begin{center}
    \includegraphics[width=1.\textwidth]{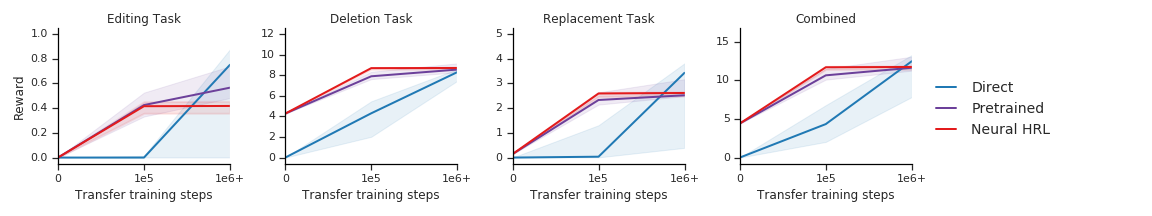}
    \caption{Rewards obtained on the editing, deletion and addition tasks, and the combined task, for the agents trained on the combinations of tasks. The lines show the median over 10 seeds, and shaded areas indicate best and worst seeds.}
    \label{fig:controller_combined_details}
\end{center}
\end{figure}

We present on \autoref{fig:controller_combined_details} the detail of the agents performance in the combined task. We note that training on the combination of tasks has positive interference which improves the performance of the agent directly trained on the tasks. Nonetheless, the hierarchical controller still produces the best performance in the low data training regime.

\end{document}